\documentclass[dvips,coll,linenumbers]{imsart}
\RequirePackage[OT1]{fontenc}
\usepackage{amsthm,amsmath,natbib}
\RequirePackage[colorlinks,citecolor=blue,urlcolor=blue]{hyperref}
\RequirePackage{hypernat}

\usepackage{amsmath,amssymb,amsfonts,amsthm}
\usepackage{bbm}
\usepackage{epsfig, psfrag}%, psfig, pstricks}
\usepackage{multirow,verbatim}

\pubyear{0000}
\volume{0}
\volumetitle{Volume Title}
\firstpage{1}
\lastpage{6}
\arxiv{math.PR/0000000}

\startlocaldefs
\numberwithin{equation}{section}
\theoremstyle{plain}

\endlocaldefs

\newtheorem{definition}{Definition}%[section]
\newcommand{\bx}{\mbox{\bf x}}

\newcommand{\bX}{\mbox{\bf X}}

\newcommand{\bzero}{\mbox{\bf 0}}

\newcommand{\bbeta}{\mbox{\boldmath $\beta$}}

\newcommand{\real}{I\kern-0.37emR}

\begin{document}

\begin{frontmatter}
\title{High-dimensional variable selection for Cox's proportional hazards model}
\runtitle{High-dimensional variable selection}
%\thankstext{T1}{Footnote to the title with the `thankstext' command.}

\begin{aug}
\author{\fnms{Jianqing} \snm{Fan}\thanksref{a,t2}\ead[label=e1]{jqfan@princeton.edu}},
\author{\fnms{Yang} \snm{Feng}\thanksref{a}\ead[label=e2]{yangfeng@princeton.edu}}
\and
\author{\fnms{Yichao} \snm{Wu}\thanksref{b, t3}%
\ead[label=e3]{wu@stat.ncsu.edu}%
\ead[label=u1,url]{http://www.foo.com}}

\thankstext{t2}{Jianqing Fan's research is partially supported by the NSF
grants DMS-0704337, DMS-0714554 and NIH grant R01-GM072611.}

\thankstext{t3}{Yichao Wu's research is partially supported by the NSF grant DMS-0905561 and the NIH/NCI grant R01-CA149569.}

\runauthor{J. Fan et al.}

\affiliation{Princeton University and North Carolina State University}

\address[a]{Jianqing Fan is Frederick L. Moore '18 Professor of Finance and Professor of Statistics, Yang Feng is Ph.D. candidate, Department of Operations Research
and Financial Engineering, Princeton University, Princeton, NJ
08544, \printead{e1,e2}}

\address[b]{Yichao Wu is Assistant Professor, Department of Statistics, North Carolina State University, Raleigh, NC 27695,
\printead{e3}}

\contributor{Fan, J.}{Princeton University}
\contributor{Feng, Y.}{Princeton University}
\contributor{Wu, Y.}{North Carolina State University}

\end{aug}

\begin{abstract}
Variable selection in high dimensional space has challenged many contemporary statistical problems from many frontiers of scientific disciplines.  Recent technological advances have made it possible to collect a huge amount of covariate information such as microarray, proteomic and SNP data via bioimaging technology while observing survival information on patients in clinical studies. Thus, the same challenge applies in survival analysis in order to understand the association between genomics information and clinical information about the survival time. In this work, we extend the sure screening procedure \citep{FanLv2008} to Cox's proportional hazards model with an iterative version available. Numerical simulation studies have shown encouraging performance of the proposed method in comparison with other techniques such as LASSO.  This demonstrates the utility and versatility of the iterative sure independence screening scheme.
\end{abstract}



\tableofcontents

\end{frontmatter}
\section{Introduction} \label{Sec:Intro}%\setzero
%\vskip-5mm \hspace{5mm }

Survival analysis is a commonly-used method for the analysis of failure time such as biological death, mechanical failure, or credit default. Within this context, death or failure is also referred to as an ``event". Survival analysis tries to model time-to-event data, which is usually subject to censoring due to the termination of study.  The main goal is to study the dependence of
the survival time $T$ on covariate variables $\bX=(X_1, X_2, \cdots,
X_p)^T$, where $p$ denotes the dimensionality of the covariate space. One common way of achieving this goal is hazard regression, which studies how the conditional hazard function of $T$ depends on the covariate $\bX=\bx$, which is defined as
$$h(t|\bx)=\lim_{\triangle t \downarrow 0} \frac{1}{\triangle t} P\{t\leq T< t+\triangle t|T\geq t, \bX=\bx\}.$$ According to the definition, the conditional hazard function is nothing but the instantaneous rate of failure at time $t$ given a particular value $\bx$ for the covariate $\bX$. The proportional hazards  model is very popular, partially due to its simplicity and its convenience in dealing with censoring. The model assumes that
\begin{equation}
h(t|\bx)=h_0(t)\Psi(\bx), \nonumber%\label{propmodel}
\end{equation}
in which $h_0(t)$ is the baseline hazard function and $\Psi(\bx)$ is the covariate effect.
Note that this model is not uniquely determined in that $ch_0(t)$ and $\Psi(\bx)/c$ give the same model for any $c>0$. Thus one identifiability condition needs to be specified.
When the identifiability condition $\Psi(\bzero)=1$ is enforced, the function $h_0(t)$, the conditional hazard function of $T$ given $\bX=\bzero$, is called the baseline hazard function.

By taking the reparametrization $\Psi(\bx)=e^{\psi(\bx)}$, \cite{Cox72, Cox75} introduced the proportional hazards model
$$h(t|\bx)=h_0(t)e^{\psi(\bx)}.$$ See \cite{KleMoe2005} and references therein for more detailed literature on Cox's proportional hazards model.

% is
%based on the assumption that the hazard function $h(y|\bx)$ of a
%subject with covariates $\bx$ satisfies
%\begin{equation}
%h(y|\bx)=h_0(y)e^{\bx^T\bbeta}.
%\end{equation}

Here the baseline hazard function $h_0(t)$ is typically completely unspecified and needs to be estimated nonparametrically. A linear model assumption, $\psi(\bx)=\bx^T\bbeta$, may be made, as is done in this paper.
Here $\bbeta=(\beta_1, \beta_2, \cdots, \beta_p)^T$ is the regression
parameter vector. While conducting survival analysis, we  not only
need to estimate $\bbeta$ but also have to estimate the baseline
hazard function $h_0(t)$ nonparametrically. Interested readers may consult
\cite{KleMoe2005} for more details.

Recent technological advances have made it possible to collect a huge amount of covariate information such as microarray, proteomic and SNP data via bioimaging technology while observing survival information on patients in clinical studies. However it is quite likely that not all available covariates are associated with the clinical outcome such as the survival time. In fact, typically a small fraction of covaraites are associated with the clinical time.  This is the notion of sparsity and consequently calls for the identification of important risk factors and at the same time quantifying their risk contributions when we analyze time-to-event data with many predictors. Mathematically, it means that we need
to identify which $\beta_j$s are nonzero and also estimate these
nonzero $\beta_j$s.

Most classical  model selection techniques have been extended from linear regression to survival analysis.  They include the best-subset selection, stepwise selection, bootstrap procedures
\citep{SSAUERBREISCHUMACHER1992}, Bayesian variable selection \citep{FSimon1998, Ibrahim1999}.
Please see references therein. Similarly, other more modern penalization approaches have been
extended as well. \cite{Tibshirani1997} applied the LASSO penalty to survival analysis.
\cite{fanli2002} considered survival analysis with the SCAD and other folded concave penalties. \cite{zhanglualasso} proposed
the adaptive LASSO penalty while studying time-to-event data. Among many other considerations is
\cite{LiDic2009}. Available theory and empirical results show that
 these penalization approaches  work well with a
moderate number of covariates.

Recently we have seen a surge of
interest in variable selection with an ultra-high dimensionality. By
ultra-high, \cite{FanLv2008} meant that the dimensionality grows
exponentially in the sample size, i.e., $\log(p)=O(n^a)$ for some
$a\in(0,1/2)$. For ultra-high dimensional linear regression,
\cite{FanLv2008} proposed sure independence screening (SIS) based on
marginal correlation ranking. Asymptotic theory is proved to show
that, with high probability, SIS keeps  all important predictor
variables with vanishing false selection rate. An important extension, iterative SIS (ISIS), was also
proposed to handle difficult cases such as when some important predictors
are marginally uncorrelated with the response. In order to deal with more complex real data,
\cite*{FanSamworthWu-09} extended SIS and ISIS to more general loss
based models such as generalized linear models, robust regression,
and classification and improved some important steps of the original ISIS. In particular, they proposed the concept of conditional marginal regression and a new variant of the method based on splitting samples. A non-asymptotic theoretical
result shows that the splitting based new variant can reduce false
discovery rate. Although the extension of \cite*{FanSamworthWu-09}
covers a wide range of statistical models, it has not been explored
whether the iterative sure independence screening method can be extended to hazard regression with censoring event time.  In this work, we will focus on Cox's proportional hazards model and extend SIS and ISIS accordingly. Other extensions of SIS include \cite{FanSong2009} and \cite*{FanFengSong2010} to generalized linear models and nonparametric additive models, in which new insights are provided via elegant mathematical results and carefully designed simulation studies.

The rest of the article is organized as follows. Section \ref{Sec:Cox} details the Cox's proportional hazards model. An overview of variable
selection via penalized approach is given in Section \ref{Sec:Variable-Selection-Cox} for Cox's proportional hazards model. We extend the SIS and ISIS procedures to Cox's model in
Section \ref{Sec:SIS-Cox}. Simulation results in Section \ref{Sec:Simulation} and real data
analysis in Section \ref{Sec:RealData} demonstrate the effectiveness of the proposed SIS and ISIS methods.
%We give a short conclusion in Section \ref{Sec:Conclusion}.

\section{Cox's proportional hazards models}\label{Sec:Cox}%\setzero
%\vskip-5mm \hspace{5mm }

 Let $T$, $C$, and $\bX$ denote the
survival time, the censoring time, and their associated covariates,
respectively. Correspondingly, denote by $Y=\min\{T, C\}$ the
observed time and $\delta=I(T\leq C)$ the censoring indicator.
For simplicity we assume that $T$ and $C$ are conditionally
independent given $\bX$ and that the censoring mechanism is
noninformative. Our observed data set $\{(\bx_i, y_i, \delta_i):
\bx_i\in \real^p, y_i\in \real^+, \delta_i\in\{0,1\}, i=1, 2,
\cdots, n\}$ is an independently and identically distributed random
sample from a certain population $(\bX, Y, \delta)$. Define
$\mathcal{C}=\{i: \delta_i=0\}$ and $\mathcal{U}=\{i: \delta_i=1\}$
to be the censored and uncensored index sets, respectively. Then the
complete likelihood of the observed data set is given by
\begin{equation}
L=\prod_{i\in\mathcal{U}}
f(y_i|\bx_i)\prod_{i\in\mathcal{C}}\bar{F}(y_i|\bx_i)=\prod_{i\in\mathcal{U}}h(y_i|\bx_i)
\prod_{i=1}^{n}\bar{F}(y_i|\bx_i), \nonumber
\end{equation}
where $f(t|\bx)$, $\bar{F}(t|\bx)=\int_t^{\infty} f(s|\bx)ds$, and $h(t|\bx)=f(t|\bx)/\bar{F}(t|\bx)$ are the
conditional density function, the conditional survival function, and
the conditional hazard function of $T$ given $\bX=\bx$, respectively.

Let $t_1^0< t_2^0 < \cdots < t_N^0$ be the ordered distinct observed
failure times. Let $(j)$ index its associate covariates $\bx_{(j)}$ and
$\mathcal{R}(t)$ be the risk set right before the time
$t$: $\mathcal{R}(t)=\{i: y_i\geq t\}$. Consider the proportional
hazards model,
\begin{equation}
h(t|\bx)=h_0(t)\exp(\bx^T\bbeta), \label{coxmodelwu}
\end{equation}
where $h_0(t)$ is the baseline hazard function. In this model, both
$h_0(t)$ and $\bbeta$ are unknown and have to be estimated. Under model \eqref{coxmodelwu}, the likelihood becomes
\begin{equation}
L=\prod_{j=1}^{N}h_0(y_{(j)})\exp(\bx_{(j)}^T\bbeta)\prod_{i=1}^{n}\exp\{-H_0(y_i)\exp(\bx_i^T\bbeta)\}, \nonumber
\end{equation}
where $H_0(t)=\int_{0}^{t}h_0(s)ds$ is the corresponding cumulative baseline hazard
function.

Following Breslow's idea, consider the ``least informative"
nonparametric modeling for $H_0(\cdot)$, in which $H_0(t)$ has a
possible jump $h_j$ at the observed failure time $t_{j}^0$, namely,
$H_0(t)=\sum_{j=1}^{N} h_j I(t_j^0\leq t)$. Then
\begin{equation}
H_0(y_i)=\sum_{j=1}^{N}h_jI(i\in
\mathcal{R}(t_j^0)).\label{leastinfo}
\end{equation}
 Consequently the
log-likelihood becomes
\begin{equation}
\sum_{j=1}^{N}
\{\log(h_j)+\bx_{(j)}^T\bbeta\}-\sum_{i=1}^{n}\{\sum_{j=1}^{N} h_j
I(i\in\mathcal{R}(t_j^0))\exp(\bx_i^T\bbeta)\}. \nonumber
\end{equation}
Maximizer $h_j$ is given by
\begin{equation}
\hat
h_j(\bbeta)=\{\sum_{i\in\mathcal{R}(t_j^0)}\exp(\bx_i^T\bbeta)\}^{-1}.\label{hj}
\end{equation}
Putting this maximizer back to the log-likelihood, we get
\begin{equation}
\sum_{j=1}^{N}[\bx_{(j)}^T\bbeta-\log\{\sum_{i\in\mathcal{R}(t_j^0)}\exp(\bx_i^T\bbeta)\}],\nonumber
\end{equation}
which is equivalent to % Using the censoring indicator $\delta_i$, it
%is equivalent to
\begin{equation}
\ell(\bbeta)=\sum_{i=1}^{n}\delta_i\bx_i^T\bbeta
-\sum_{i=1}^{n}\delta_i \log\{\sum_{j\in\mathcal{R}(y_i)}
\exp(\bx_j^T\bbeta)\}. \label{simlik}
\end{equation}
by using the censoring indicator $\delta_i$.  This is the partial likelihood due to \cite{Cox75}.

Maximizing $\ell(\bbeta)$ in (\ref{simlik}) with respect to
$\bbeta$, we can get an estimate $\hat\bbeta$ of the regression
parameter. Once $\hat\bbeta$ is available, we may plug it into
(\ref{hj}) to get $\hat h_j(\hat\bbeta)$. These newly obtained $\hat
h_j(\hat\bbeta)$s can be plugged into (\ref{leastinfo}) to obtain
our nonparametric estimate of the baseline cumulative hazard
function.

\section{Variable selection for Cox's proportional hazards model via
penalization}\label{Sec:Variable-Selection-Cox}

In the estimation scheme presented in the previous section, none of the estimated regression coefficients is exactly zero,
leaving all covariates in the final model. Consequently it is incapable of selecting important variables and handling the case with $p > n$. To achieve variable
selection, classical  techniques such as the best-subset selection,
stepwise selection, and bootstrap procedures have been extended
accordingly to handle Cox's proportional hazards model.

In this section, we will focus on some more advanced  techniques for
variable selection  via penalization. Variable selection via
penalization has received lots of attention recently. Basically it
uses some variable selection-capable penalty function to regularize
the objective function while performing optimization. Many variable
selection-capable penalty functions have been proposed. A well known
example is  the $L_1$ penalty  $\lambda\sum_{j=1}^{p}|\beta_j|$,
which is also known as the LASSO penalty \citep{tibshirani1996}.
Among many others are the SCAD penalty \citep{fanli2001},
elastic-net penalty \citep{ZouHastie2005}, adaptive $L_1$
\citep{zoualasso, zhanglualasso}, and minimax concave penalty
\citep{zhangch2009}.

Denote a general penalty function by $p_{\lambda}(\cdot)$, where
$\lambda>0$ is a regularization parameter. From derivations in the last section, penalized likelihood is equivalent to penalized partial likelihood:
While maximizing
$\ell(\bbeta)$ in (\ref{simlik}), one may regularize it using
$\sum_{j=1}^{p}p_{\lambda}(\beta_j)$. Equivalently we solve
\begin{equation}
\min -\ell(\bbeta)+\sum_{j=1}^{p}p_{\lambda}(\beta_j) \nonumber
\end{equation}
by including a negative sign in front of $\ell(\bbeta)$. In the
literature,  \cite{Tibshirani1997}, \cite{fanli2002}, and
\cite{zhanglualasso} considered the $L_1$, SCAD, and adaptive $L_1$
penalties while studying time-to-event data,  respectively,  among many others.

\begin{figure}[h!]
\centering{\scalebox{0.6}{\includegraphics{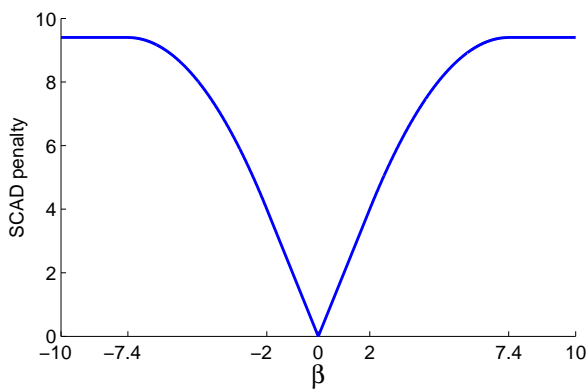}}}
\caption{Plot of the SCAD penalty with $a=3.7$ and $\lambda=2$}\label{scadplot}
\end{figure}

In this paper, we will use the SCAD penalty  for our  extensions of
SIS and ISIS whenever necessary. The SCAD function is a quadratic
spline and symmetric around the origin. It can be defined in terms
of its first order derivative
\[
p'_\lambda(|\beta|)=\lambda\biggl\{\mathbbm{1}_{\{|\beta|\leq\lambda\}}
+\frac{(a\lambda-|\beta|)_+}{(a-1)\lambda}
\mathbbm{1}_{\{|\beta|>\lambda\}}\biggr\},
\]
for some $a > 2$ and $\beta \neq 0$. Here $a$ is a parameter and
\cite{fanli2001} recommend to use $a=3.7$ based on a Bayesian
argument. The SCAD penalty is plotted in Figure \ref{scadplot} for $a=3.7$ and $\lambda=2$.
The SCAD penalty is non-convex, leading to non-convex optimization. For the non-convex SCAD penalized optimization, \cite{fanli2001} proposed the local quadratic approximation; \cite{ZouLi08} proposed the local linear approximation; \cite{WuLiu09} presented the difference convex algorithm. In this work, whenever necessary we use the local linear approximation algorithm to solve the SCAD penalized optimization. 
%In each iteration of the local linear approximation, we actually use the weighted LASSO penalties with the weights calculated using the SCAD penalty. The local linear approximation can also be used to solve the adaptive LASSO penalized problem, which bridges the two penalties as specific members of the folded concave penalized likelihood in \cite{fanli2001}.

%appropriate penalty such as the LASSO and SCAD will be applied to
%achieve variable selection. With the scad penalty
%$p_\lambda(\cdot)$, we are minimizing
%\begin{equation}
%-\sum_{i=1}^{n}\delta_i\bx_i^T\bbeta +\sum_{i=1}^{n}\delta_i
%\log(\sum_{j\in\mathcal{R}(y_i)}
%\exp(\bx_j^T\bbeta))+\sum_{j=1}^{p}p_\lambda(\beta_j).
%\end{equation}
%
%Denote $\ell(\bbeta)=-\sum_{i=1}^{n}\delta_i\bx_i^T\bbeta
%+\sum_{i=1}^{n}\delta_i \log(\sum_{j\in\mathcal{R}(y_i)}
%\exp(\bx_j^T\bbeta))$.
%
%Then the generalized ISIS algorithm can be applied.

\section{SIS and ISIS for Cox's proportional hazard model}\label{Sec:SIS-Cox}

The penalty based variable selection techniques work great with a
moderate number of covariates. However their usefulness is limited
while dealing with an ultra-high dimensionality as shown in
\cite{FanLv2008}. In the linear regression case, \cite{FanLv2008}
proposed to rank covariates according to the absolute value of their
marginal correlation with the response variable and select the top
ranked covariates. They provided theoretical result to guarantee
that this simple correlation ranking retains all important covariates
with high probability. Thus they named their method sure
independence screening (SIS). In order to handle difficult problems
such as the one with some important covariates being marginally uncorrelated
with the response, they proposed iterative SIS (ISIS). ISIS begins
with SIS, then it regresses the response on covariates selected by
SIS and uses the regression residual as a  ``working" response  to
recruit more covariates with SIS. This process can be repeated until
some convergence criterion has been met. Empirical improvement over
SIS has been observed for ISIS. In order to increase the power of the sure independence screening technique,
\cite*{FanSamworthWu-09} has extended
SIS and ISIS to more general models such as generalized linear
models, robust regression, and classification and made several important improvements.  We now extend the key idea of SIS and ISIS to handle Cox's proportional hazards model.

Let $\mathcal{M}^*$ be the index set of the true underlying
sparse model, namely, $\mathcal{M}^*=\{j: \beta_j^*\ne0\mbox{ and }
1\leq j \leq p\}$,  where $\beta_j^*$s are the true regression
coefficients in the Cox's proportional hazards model (\ref{coxmodelwu}).

\subsection{Ranking by marginal utility}
First let us review the definition of sure screening property.
\begin{definition}[Sure Screening Property] We say a model selection procedure satisfies sure screening property if the selected model $\hat{\mathcal{M}}$ with model size $o_p(n)$ includes the true model $\mathcal{M}^*$ with probability tending to one.
\end{definition}
For each covariate $X_m$ ($1\leq m\leq p$), define its marginal
utility as the maximum of the partial likelihood of the single covariate:
\begin{equation}
u_m=\max_{\beta_m} \left(\sum_{i=1}^{n}\delta_i x_{im}\beta_m
-\sum_{i=1}^{n}\delta_i \log\{\sum_{j\in\mathcal{R}(y_i)}
\exp(x_{jm}\beta_m)\}\right). \nonumber
\end{equation}
Here $x_{im}$ is the $m$-th element of $\bx_i$, namely $\bx_i=(x_{i1}, x_{i2}, \cdots, x_{ip})^T$.
Intuitively speaking, the larger the marginal utility is, the more
information the corresponding covariate contains the information about
the survival outcome. Once we have obtained all marginal utilities
$u_m$ for $m=1, 2, \cdots, p$, we rank all covariates according to
their corresponding marginal utilities from the largest to the
smallest and select the $d$ top ranked covariates. Denote by
$\mathcal{I}$ the index set of these $d$ covariates that have
been selected.

The index set $\mathcal{I}$ is expected to cover the true index set
$\mathcal{M}^*$ with a high probability, especially when we use a
relative large $d$. This is formally shown by \cite{FanLv2008} for the linear model with Gaussian noise and Gaussian covariates and significantly expanded by \cite{FanSong2009} to generalized linear models with non-Gaussian covariates.  The parameter $d$ is usually chosen large enough to ensure the sure screening property.  However the estimated index set $\mathcal{I}$  may also include a lot of unimportant covariates. To improve performance, the penalization based variable selection approach can be applied to the selected subset of the variables $\{X_j, j \in \mathcal{I} \}$ to further delete unimportant variables.  Mathematically, we then solve the following penalized partial likelihood problem:
\begin{equation}
\min_{\bbeta_{\mathcal{I}}}\left(
-\sum_{i=1}^{n}\delta_i\bx_{\mathcal{I},i}^T\bbeta_{\mathcal{I}}
+\sum_{i=1}^{n}\delta_i \log\{\sum_{j\in\mathcal{R}(y_i)}
\exp(\bx_{\mathcal{I},j}^T\bbeta_{\mathcal{I}})\}+\sum_{m \in
\mathcal{I}}p_{\lambda}(\beta_m)\right), \nonumber
\end{equation}
where $\bx_{\mathcal{I},i}$ denotes a sub-vector of $\bx_i$ with
indices in $\mathcal{I}$ and similarly for $\bbeta_{\mathcal{I}}$.
It will lead to sparse regression parameter estimate $\hat
\bbeta_{\mathcal{I}}$. Denote the index set of nonzero components of
$\hat \bbeta_{\mathcal{I}}$ by $\hat{\mathcal{M}}$, which will serve
as our final estimate of $\mathcal{M}^*$.

\subsection{Conditional feature ranking and iterative feature selection}

\cite{FanLv2008} pointed out that SIS can fail badly for some challenging scenarios such as the case that there exist
jointly related but marginally unrelated covariates or jointly
uncorrelated covariates having higher marginal correlation with the
response than some important predictors. To deal with such difficult
scenarios, iterative SIS (ISIS) has been proposed. Comparing to SIS
which is based on marginal information only, ISIS tries to make more
use of joint covariates' information.

The iterative SIS begins with using  SIS to select an index set
$\hat{\mathcal{I}}_1$, upon which a penalization based variable selection
step is applied to get regression parameter estimate
$\hat\bbeta_{\hat{\mathcal{I}}_1}$.  A refined estimate
 of the true index set is obtained and denoted by $\hat{\mathcal{M}}_1$,  the index set corresponding to nonzero elements of $\hat\bbeta_{\hat{\mathcal{I}}_1}$.

As in \citet*{FanSamworthWu-09}, we next define the conditional utility of
each covariate $m$ that is not in $\hat{\mathcal{M}}_1$ as follows:
\begin{eqnarray*}
u_{m|\hat{\mathcal{M}}_1}&=&\max_{\beta_m,
\bbeta_{\hat{\mathcal{M}}_1}} \left(\left.\sum_{i=1}^{n}\delta_i
(x_{im}\beta_m+\bx_{\hat{\mathcal{M}}_1,
i}^T\bbeta_{\hat{\mathcal{M}}_1})\right. \right.\\
&&\left.-\sum_{i=1}^{n}\delta_i
\log\{\sum_{j\in\mathcal{R}(y_i)} \exp(x_{jm}^T\beta_m+
\bx_{\hat{\mathcal{M}}_1, j}^T\bbeta_{\hat{\mathcal{M}}_1}
)\}\right).\nonumber
\end{eqnarray*}
%$j\not\in \hat{\mathcal{M}}_1$.
This conditional utility measures the additional contribution of the
$m$th covariate given that all covariates with indices in
$\hat{\mathcal{M}}_1$ have been included in the model.

Once the conditional utilities have been defined for each covariate
that is not in $\hat{\mathcal{M}}_1$, we rank them from the largest
to the smallest and select these covariates with top rankings.
Denote the index set of these selected covariates by
$\hat{\mathcal{I}}_2$. With $\hat{\mathcal{I}}_2$ having been identified, we
minimize
\begin{eqnarray}
& & -\sum_{i=1}^{n}\delta_i (\bx_{\hat{\mathcal{M}}_1\cup \hat{\mathcal{I}}_2,
i}^T\bbeta_{\hat{\mathcal{M}}_1\cup \hat{\mathcal{I}}_2})
+\sum_{i=1}^{n}\delta_i \log\{\sum_{j\in\mathcal{R}(y_i)}
\exp(\bx_{\hat{\mathcal{M}}_1\cup \hat{\mathcal{I}}_2,
j}^T\bbeta_{\hat{\mathcal{M}}_1\cup \hat{\mathcal{I}}_2}
)\}\\
& & \hspace*{0.3 in} +\sum_{m\in \hat{\mathcal{M}}_1\cup \hat{\mathcal{I}}_2}
p_{\lambda}(\beta_j)\nonumber
\end{eqnarray}
with respect to $\bbeta_{\hat{\mathcal{M}}_1\cup
\hat{\mathcal{I}}_2}$ to get sparse estimate
$\hat\bbeta_{\hat{\mathcal{M}}_1\cup \hat{\mathcal{I}}_2}$. Denote
the index set corresponding to nonzero components of
$\hat\bbeta_{\hat{\mathcal{M}}_1\cup \hat{\mathcal{I}}_2}$ to be
$\hat{\mathcal{M}}_2$, which is our updated estimate of the true
index set $\mathcal{M}^*$.  Note that this step can delete some variables $\{X_j \in \hat{\mathcal{M}}_1\}$ that were previously selected.  This idea was proposed in \cite{FanSamworthWu-09} and is an improvement of the idea in \cite{FanLv2008}.

The above iteration can be repeated until some convergence criterion
is reached. We adopt the criterion of either having identified $d$
covariates or $\hat{\mathcal{M}}_j=\hat{\mathcal{M}}_{j-1}$ for some
$j$.

\subsection{New variants of SIS and ISIS for reducing FSR}
\cite*{FanSamworthWu-09} noted that the idea of sample spliting
can also be used to reduce the false selection rate.
Without loss of generality, we assume that the sample size
$n$ is even. We randomly split the sample into two halves. Then
apply SIS or ISIS separately to the data in each partition to obtain
two estimates $\hat{\mathcal{I}}^{(1)}$ and
$\hat{\mathcal{I}}^{(2)}$ of the true index set
$\mathcal{M}^{*}$. Both these two estimates could have high
FSRs because they are based on a simple and crude screening method.
Yet each of them should include all important covariates with high
probabilities. Namely, important covariates should appear in both
sets with probability tending to one asymptotically. Define a new
estimate by intersection $\hat{\mathcal{I}}=\hat{\mathcal{I}}^{(1)}
\cap \hat{\mathcal{I}}^{(2)}$. The new estimate $\hat{\mathcal{I}}$
should include all important covaraites with high probability as well due
to properties of each individual estimate. However by construction,
the number of unimportant covariates in the new estimate
$\hat{\mathcal{I}}$ is much smaller. The reason is that, in order
for an unimportant covariate to appear in $\hat{\mathcal{I}}$, it
has to be included in both $\hat{\mathcal{I}}^{(1)}$ and
$\hat{\mathcal{I}}^{(2)}$ randomly.

For the new variant method based on random splitting, \cite*{FanSamworthWu-09} obtained some non-asymptotic probability
bound for  the event that  $r$ unimportant covaraites are included
in the intersection $\hat{\mathcal{I}}$ for any natural number $r$
under some exchangeability condition on all unimportant covariates.
The probability bound is decreasing in the dimensionality, showing a
``blessing of dimensionality". Please consult
\cite*{FanSamworthWu-09} for more details. We want to remark that
their theoretical bound is applicable to our setting as well while studying
time-to-event data because theoretical bound is based on splitting the sample into two halves and only requires the independence between these two halves.

While defining new variants, we may use the same $d$ as used in the
original SIS and ISIS. However it will lead to a very aggressive
screening. We call the corresponding variant the first variant of
(I)SIS. Alternatively, in each step we may choose larger
$\hat{\mathcal{I}}^{(1)}$ and $\hat{\mathcal{I}}^{(2)}$ to ensure
that their intersection $\hat{\mathcal{I}}^{(1)}\cap
\hat{\mathcal{I}}^{(2)}$ has $d$ covariates, which is called the
second variant. The second variant ensures that there are at least
$d$ covariates included before applying penalization in each step
and is thus  much less aggressive. Numerical examples will be used
to explore their performance and prefer to the first variant.

%\section{Sure screening property}
%Yang: do you have any idea whether we can extend Fan and Song
%(2009)'s proof to establish sure screening property to our case? If
%so, that will improve our paper significantly.

\section{Simulation}\label{Sec:Simulation}
\subsection{Design of simulations}
In this section, we conduct simulation studies to show the power of the (I)SIS and its variants by
comparing them with LASSO \citep{Tibshirani1997} in the Cox's proportional hazards model. Here the regularization parameter for LASSO is tuned via five fold cross validation. Most of the settings are adapted from \cite{FanLv2008} and \cite*{FanSamworthWu-09}. Four different
configurations are considered with $n=300$ and $p=400$. And two of them are revisited with a
different pair of sample size $n=400$ and  dimensionality  $p=1000$. Covariates in different settings are generated  as follows.
\begin{itemize}
\item[Case] 1: $X_1, \cdots, X_p$ are independent and identically distributed  $N(0,1)$ random variables.
\item[Case] 2: $X_1, \cdots, X_p$ are multivariate Gaussian, marginally $N(0,1)$, and with serial correlation $\mbox{corr}(X_i, X_j)=\rho$ if $i\neq j$. Here we take $\rho=0.5$.
\item[Case] 3: $X_1, \cdots, X_p$ are multivariate  Gaussian, marginally $N(0,1)$, and with correlation structure $\mbox{corr}(X_i,X_4)=1/\sqrt{2}$ for all $i\neq 4$ and $\mbox{corr}(X_i,X_j)=1/2$ if $i$ and $j$
are distinct elements of $\{1,\cdots, p\} \backslash \{4\}$.
\item[Case] 4: $X_1, \cdots, X_p$ are multivariate Gaussian, marginally $N(0,1)$, and with correlation structure $\mbox{corr}(X_i,X_5)=0$ for all $i\neq 5$, $\mbox{corr}(X_i,X_4)=1/\sqrt{2}$ for all $i\notin \{4,5\}$, and $\mbox{corr}(X_i, X_j)=1/2$ if $i$ and $j$ are distinct elements of $\{1,\cdots,p\}\backslash \{4,5\}$.
\item[Case] 5: Same as Case 2 except $n=400$ and $p=1000$.
\item[Case] 6: Same as Case 4 except $n=400$ and $p=1000$.
\end{itemize}

Here, Case 1 with independent predictors is the most straightforward for
variable selection. In Cases 2-6, however, we have serial
correlation such that $\mbox{corr}(X_i,X_j)$ does not decay as
$|i-j|$ increases. We will see later that for Cases 3, 4 and 6, the
true coefficients are specially chosen such that the response is marginally
independent but jointly dependent of $X_4$. We therefore expect
variable selection in these situations to be much more challenging,
especially for the non-iterated versions of SIS. Notice that in the
asymptotic theory of SIS in \cite{FanLv2008}, this type of
dependence is ruled out by their Condition (4).

%As suggested in \citet{FanSamworthWu-09},

In our implementation, we choose $d=\lfloor \frac{n}{4\log n}
\rfloor$ for both the vanilla version of SIS (Van-SIS) and the
second variant (Var2-SIS). For the first variant (Var1-SIS),
however, we use $d=\lfloor \frac{n}{\log n} \rfloor$ (note that
since  the selected variables for the first variant are in the intersection of
two sets of size $d$, we typically end up with far fewer than $d$
variables selected by this method). For any type of SIS or ISIS, we apply SCAD with these selected predictors to get a final estimate of the regression coefficients at the end of the screening step. Whenever necessary, the BIC is used to select the best tuning parameter in the regularization framework.

In all setting, the censoring time is generated from exponential distribution with mean $10$. This corresponds to choosing the baseline hazard function $h_0(t)=0.1$ for $t\geq 0$. The true regression coefficients and censoring rate in each of the six cases  are as follows:
\begin{itemize}
\item[Case] 1: $\beta_1=-1.6328, \beta_2=1.3988, \beta_3=-1.6497, \beta_4=1.6353, \beta_5=-1.4209, \beta_6=1.7022$, and $\beta_j=0$ for $j>6$. The corresponding censoring rate is 33\%.
\item[Case] 2: The coefficients are the same as Case 1. The
corresponding censoring rate is 27\%.
\item[Case] 3: $\beta_1=4, \beta_2=4, \beta_3=4, \beta_4=-6\sqrt{2}$, and $\beta_j=0$ for $j>4$. The
corresponding censoring rate is 30\%.
\item[Case] 4: $\beta_1=4, \beta_2=4, \beta_3=4, \beta_4=-6\sqrt{2}, \beta_5=4/3$ and $\beta_j=0$ for $j>5$. The
corresponding censoring rate is 31\%.
\item[Case] 5: $\beta_1=-1.5140, \beta_2=1.2799, \beta_3=-1.5307, \beta_4=1.5164, \beta_5=-1.3020, \beta_6=1.5833$, and $\beta_j=0$ for $j>6$.
The corresponding censoring rate is 23\%.
\item[Case] 6: The coefficients are the same as Case 4. The corresponding censoring rate is 36\%.
\end{itemize}
In Cases 1, 2 and 5 the coefficients were chosen randomly, and were
generated as $(4 \log n/\sqrt{n}+ |Z|/4)U$ with $Z\sim N(0, 1)$ and
$U = 1$ with probability 0.5 and -1 with probability 0.5,
independent of $Z$. For Cases 3, 4, and 6, the choices ensure that
even though $\beta_4\neq 0$, we have that $X_4$ and $Y$ are
marginally independent. The fact that $X_4$ is marginally
independent of the response is designed to make it difficult for the
common independent learning to select this variable. In Cases 4 and
6, we add another important variable $X_5$ with a small coefficient
to make it even more difficult.

\subsection{Results of simulations}
%In the tables below,
We  report our simulation results based on 100 Monte Carlo repetitions for each setting in Tables 1-7.
To present our simulation results, we use several different performance measures. % all of
%which are based on 100 Monte Carlo repetitions.
In the rows labeled $\|\bbeta-\hat\bbeta\|_1$ and $\|\bbeta-\hat\bbeta\|_2^2$, we report  the median $L_1$ and squared $L_2$ estimation errors
$\|\bbeta-\hat\bbeta\|_1=\sum_{j=1}^p|\hat\beta_j-\beta_j|$ and
$\|\bbeta-\hat\bbeta\|_2^2=\sum_{j=1}^p|\hat\beta_j-\beta_j|^2$, respectively, where the median is over the 100 repetitions. In the row with label $P_1$, we report the proportion of the 100 repetitions that the (I)SIS procedure
under consideration includes all of the important variables in the
model, while the row with label $P_2$ reports the corresponding proportion of
times that the final variables selected, after further application
of the SCAD penalty, include all of the important ones. We also report the median model size (MMS) of the final model  among 100 repetitions in the row labeled MMS.

\begin{table}[h!]
\caption{Results for Cases 1 and 2. Here $P_1$ stands for the probability
that (I)SIS includes the true model, i.e., has the sure screening
property. $P_2$ stands for the probability that the final model has
the sure screening property. MMS stands for Median Model Size among
100 repetitions. The sample size $n = 300$ and the number of covariates is $p = 400$.}
\begin{center}
\begin{tabular}{c|c|c|c|c|c|c|c}
\hline
\multicolumn{8}{c}{\bf Case 1: independent covariates}\\ \hline
&\mbox{Van-SIS}&\mbox{Van-ISIS}&\mbox{Var1-SIS}&\mbox{Var1-ISIS}&\mbox{Var2-SIS}&\mbox{Var2-ISIS} & \mbox{LASSO} \\
\hline
%$\|\bbeta-\hat\bbeta\|_1$& 2.31& 1.37& 4& 1.22& 4.63& 1.19& 5.42 \\
%$\|\bbeta-\hat\bbeta\|_2^2$& 1.98& 0.49& 4.35& 0.42& 5.4& 0.38& 2.08 \\
%$P_1$& 0.78& 1& 0.5& 1& 0.39& 1& -- \\
%$P_2$& 0.78& 1& 0.5& 1& 0.39& 1& 1 \\
%MMS& 6& 6& 6& 6& 6& 6& 47 \\
$\|\bbeta-\hat\bbeta\|_1$& 0.79& 0.57& 0.73& 0.61& 0.76& 0.62& 4.23 \\
$\|\bbeta-\hat\bbeta\|_2^2$& 0.13& 0.09& 0.15& 0.1& 0.15& 0.1& 0.98 \\
$P_1$& 1& 1& 0.99& 1& 0.99& 1& -- \\
$P_2$& 1& 1& 0.99& 1& 0.99& 1& 1 \\
MMS& 7& 6& 6& 6& 6& 6& 68.5 \\ \hline
\multicolumn{8}{c}{\bf Case 2:  Equi-correlated covariates with $\rho = 0.5$}\\ \hline
&\mbox{Van-SIS}&\mbox{Van-ISIS}&\mbox{Var1-SIS}&\mbox{Var1-ISIS}&\mbox{Var2-SIS}&\mbox{Var2-ISIS} & \mbox{LASSO} \\ \hline
$\|\bbeta-\hat\bbeta\|_1$& 2.2& 0.64& 4.22& 0.8& 3.95& 0.78& 4.38 \\
$\|\bbeta-\hat\bbeta\|_2^2$& 1.74& 0.11& 4.71& 0.29& 4.07& 0.28& 0.98 \\
$P_1$& 0.71& 1& 0.42& 0.99& 0.46& 0.99& -- \\
$P_2$& 0.71& 1& 0.42& 0.99& 0.46& 0.99& 1 \\
MMS& 7& 6& 6& 6& 7& 6& 57 \\

 \hline
\end{tabular}
\label{case1}
\end{center}
\end{table}

We report results of Cases 1 and 2 in Table 1. Recall that the
covariates in Case 1 are all independent. In this case, the Van-SIS
performs reasonably well. Yet, it does not perform well for the
dependent case, Case 2. Note the only difference between Case 1 and
Case 2 is the covariance structure of the covariates. For both
cases, vanilla-ISIS and its second variant perform very well. It is
worth noticing that the ISIS improves significantly over SIS, when
covariates are dependent, in terms of both the probability of
including all the true variables and in reducing the estimation

error. This comparison indicates that the ISIS performs much better
when there is serious correlation among covariates.

While implementing the LASSO penalized Cox's proportional hazards model, we adapted the Fortran source code in the R package ``glmpath." Recall that the objective function in the LASSO penalized Cox's proportional hazards model is convex and nonlinear. What the Fortran code does is to call a MINOS subroutine to solve the corresponding nonlinear convex optimization problem. Here MINOS is an optimization software developed by Systems Optimization Laboratory at Stanford University. This nonlinear convex optimization problem is much more complicated than a general quadratic programming problem. Thus generally it takes much longer time to solve, especially so when the dimensionality is high as confirmed by Table 3. However the algorithm we used does converge as the objective function is strictly convex.

Table 1 shows that LASSO has the
sure screening property as the ISIS, however, the median model size
is ten times as large as that of ISIS. As a consequence, it also has larger
estimation errors in terms of $\|\bbeta-\hat\bbeta\|_1$ and
$\|\bbeta-\hat\bbeta\|_2^2$.  The fact that the median absolute deviation error is much larger than the median square error indicates that the LASSO selects many small nonzero coefficients for those unimportant variables.  This is also verified by the fact that LASSO has a very large median model size.  The explanation is the bias issue noted by \cite{fanli2001}.  In order for LASSO to have a small bias for nonzero coefficients, a smaller $\lambda$ should be chosen.  Yet, a small $\lambda$ recruits many small coefficients for unimportant variables. For Case 2, the LASSO has a similar performance as in Case 1 in that it includes all the important variables but has a much larger model size.

\begin{table}[h!]
\caption{Results for Cases 3 and 4.  The same caption as Table 1 is used.}
\begin{center}
\begin{tabular}{c|c|c|c|c|c|c|c}
\hline
\multicolumn{8}{c}{\bf Case 3: An important predictor that is independent of survival time}\\ \hline
&\mbox{Van-SIS}&\mbox{Van-ISIS}&\mbox{Var1-SIS}&\mbox{Var1-ISIS}&\mbox{Var2-SIS}&\mbox{Var2-ISIS} & \mbox{LASSO} \\ \hline
%$\|\bbeta-\hat\bbeta\|_1$& 20& 3.87& 20.01& 3.86& 20.04& 5.25& 20.59 \\
%$\|\bbeta-\hat\bbeta\|_2^2$& 103.93& 13.21& 106.14& 14.48& 104.71& 22.18& 78.31 \\
%$P_1$& 0& 0.89& 0& 0.89& 0& 0.81& --\\
%$P_2$& 0& 0.89& 0& 0.89& 0& 0.81& 0.15 \\
%MMS& 7& 4& 6& 4& 7& 4& 67.5 \\
$\|\bbeta-\hat\bbeta\|_1$& 20.1& 1.03& 20.01& 0.99& 20.09& 1.08& 20.53 \\
$\|\bbeta-\hat\bbeta\|_2^2$& 94.72& 0.49& 100.42& 0.47& 94.77& 0.55& 76.31 \\
$P_1$& 0& 1& 0& 1& 0& 1& -- \\
$P_2$& 0& 1& 0& 1& 0& 1& 0.06 \\
MMS& 13& 4& 8& 4& 13& 4& 118.5 \\
\hline
\multicolumn{8}{c}{\bf Case 4:  Two very hard variables to be selected.}\\ \hline
&\mbox{Van-SIS}&\mbox{Van-ISIS}&\mbox{Var1-SIS}&\mbox{Var1-ISIS}&\mbox{Var2-SIS}&\mbox{Var2-ISIS} & \mbox{LASSO} \\ \hline
$\|\bbeta-\hat\bbeta\|_1$& 20.87& 1.15& 20.95& 1.4& 20.96& 1.41& 21.04 \\
$\|\bbeta-\hat\bbeta\|_2^2$& 96.46& 0.51& 102.14& 1.77& 97.15& 1.78& 77.03 \\
$P_1$& 0& 1& 0& 0.99& 0& 0.99& -- \\
$P_2$& 0& 1& 0& 0.99& 0& 0.99& 0.02 \\
MMS& 13& 5& 9& 5& 13& 5& 118 \\
\hline
\end{tabular}
\label{case3}
\end{center}
\end{table}

Results of Cases 3 and 4 are reported in Table 2. Note that, in both cases, the design ensures that $X_4$ is
marginally independent of but jointly dependent on $Y$. This special design disables the SIS to include $X_4$ in the corresponding identified model as confirmed by our numerical results. However, by using ISIS, we are
able to select $X_4$ for each repetition. Surprisingly, LASSO rarely
includes $X_4$ even if it is not a marginal screening based method.  Case 5 is even more challenging.  In addition to the same challenge as case 4, the coefficient $\beta_5$ is 3 times smaller than the first four variables.  Through the correlation with the first 4 variables, unimportant variables $\{X_j, j \geq 6\}$ have a larger marginal utility with $Y$ than $X_5$.  Nevertheless, the ISIS works very well and demonstrates once more that it uses adequately the joint covariate information.

\begin{table}[h!]
\caption{The average running time (in seconds) comparison for
Van-ISIS and LASSO.}
\begin{center}
  \begin{tabular}{c|c|c|c|c}
  \hline
  &Case 1&Case 2&Case 3&Case 4\\
  \hline
%    Van-ISIS& 104.98& 252.32& 196.21& 177.42 \\
%LASSO& 5213.97& 6372.09& 6536.23& 6504.47 \\
Van-ISIS& 379.29& 213.44& 402.94& 231.68 \\
LASSO& 37730.82& 26348.12& 46847& 28157.71 \\
\hline
  \end{tabular}
\end{center}
\end{table}

We also compare the computational cost of van-ISIS and LASSO in
Table 3 for Cases 1-4. Table 3 shows that it takes LASSO several hours for each repetition, while van-ISIS can finish it in just several minutes. This is a huge improvement. For this reason, for Cases 5 and 6 where $p=1000$,
we only report the results for ISIS since it takes LASSO over
several days to complete a single repetition. Results of Cases 5 and 6 are reported in Table 4. The table demonstrates
similar performance as Cases 2 and 4 even with more covariates.

\begin{table}[h!]
\caption{Results for Cases 5 and 6.  The same caption as that of Table 1 is used.}
\begin{center}
\begin{tabular}{c|c|c|c|c|c|c}
\hline
\multicolumn{7}{c}{\bf Case 5: The same as case 2 with $p = 1000$ and $n=400$}\\ \hline
&\mbox{Van-SIS}&\mbox{Van-ISIS}&\mbox{Var1-SIS}&\mbox{Var1-ISIS}&\mbox{Var2-SIS}&\mbox{Var2-ISIS}  \\ \hline
$\|\bbeta-\hat\bbeta\|_1$& 1.53& 0.52& 3.55& 0.55& 2.95& 0.51 \\
$\|\bbeta-\hat\bbeta\|_2^2$& 0.9& 0.07& 3.48& 0.08& 2.5& 0.07 \\
$P_1$& 0.82& 1& 0.39& 1& 0.5& 1 \\
$P_2$& 0.82& 1& 0.39& 1& 0.5& 1 \\
MMS& 8& 6& 6& 6& 7& 6 \\
\hline
\multicolumn{7}{c}{\bf Case 6: The same as case 4 with $p=1000$ and $n=400$.}\\ \hline
&\mbox{Van-SIS}&\mbox{Van-ISIS}&\mbox{Var1-SIS}&\mbox{Var1-ISIS}&\mbox{Var2-SIS}&\mbox{Var2-ISIS}  \\ \hline
$\|\bbeta-\hat\bbeta\|_1$& 20.88& 0.99& 20.94& 1.1& 20.94& 1.29 \\
$\|\bbeta-\hat\bbeta\|_2^2$& 93.53& 0.39& 104.76& 0.44& 94.02& 1.35 \\
$P_1$& 0& 1& 0& 1& 0& 0.99 \\
$P_2$& 0& 1& 0& 1& 0& 0.99 \\
MMS& 16& 5& 8& 5& 16& 5 \\
\hline
\end{tabular}
\label{case5}
\end{center}
\end{table}

\begin{figure}[h!]
\centering{\rotatebox{270}{\includegraphics[scale=0.5]{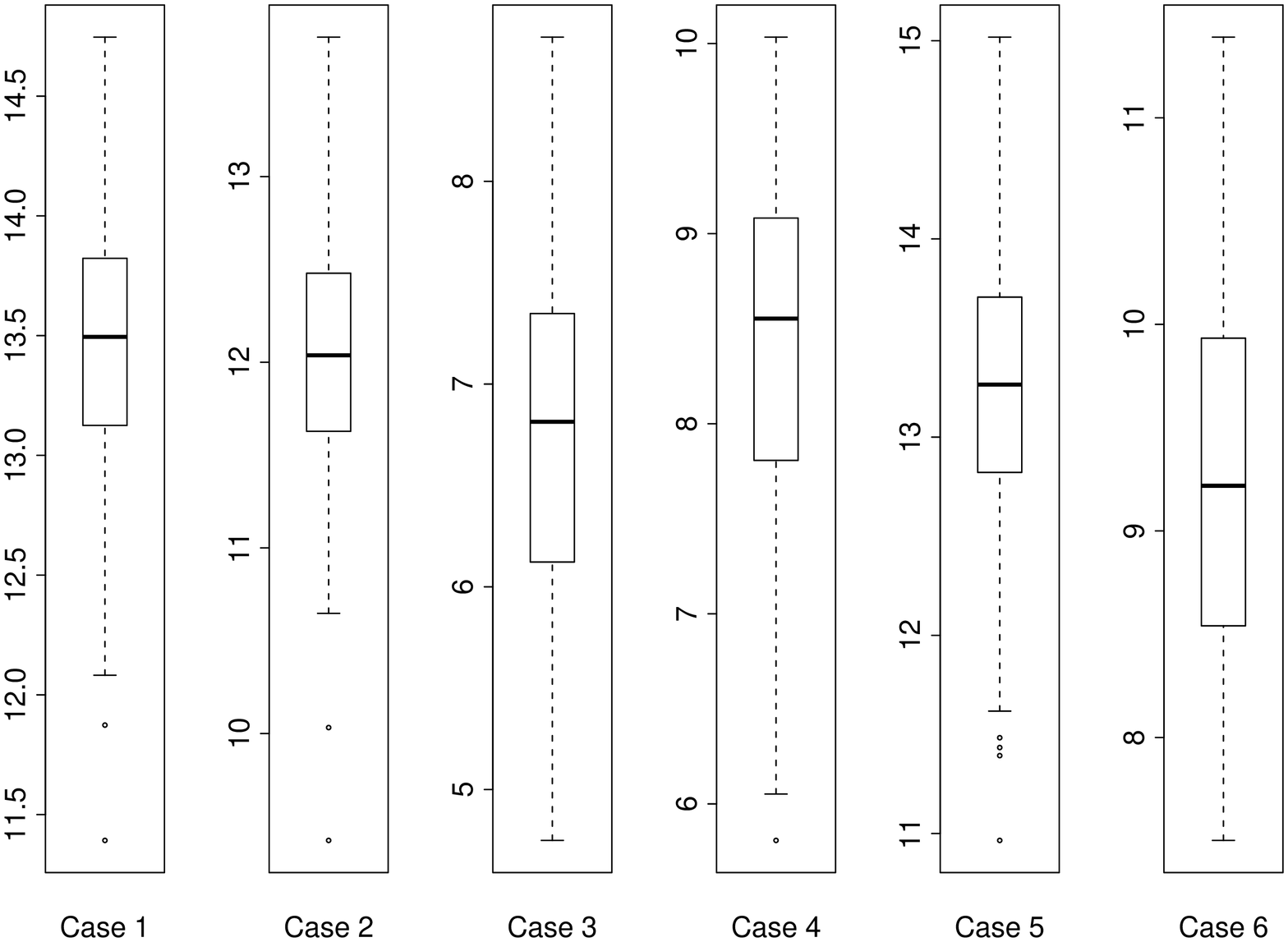}}}
\caption{The boxplots of the minimum $|t|$-statistic in the oracle
models among 100 simulations.}\label{minz}
\end{figure}

\begin{figure}[h!]
\centering{\rotatebox{270}{\includegraphics[scale=0.5]{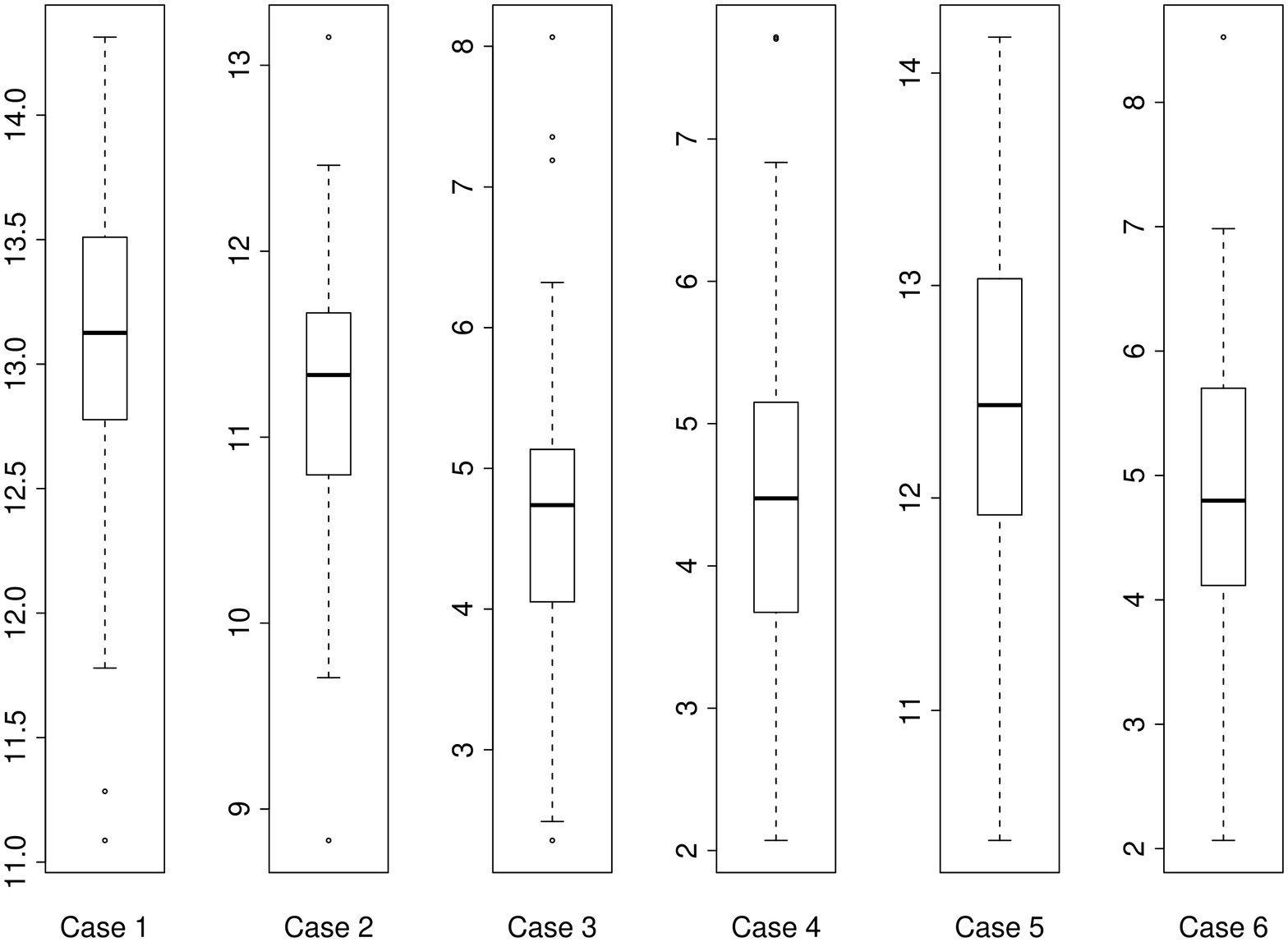}}}
\caption{The boxplots of the minimum $|t|$-statistic in the models
where 20 noise variables are added among 100
simulations.}\label{minzn}
\end{figure}

To conclude the simulation section, we demonstrate the difficulty of our simulated models by  showing the
distribution, among 100 simulations, of the minimum $|t|$-statistic
for the estimates of the true nonzero  regression coefficients in the oracle model with only true important predictors included. More explicitly, during each repetition of each simulation setting, we pretend to know the index set $\mathcal{M}^*$ the true underlying sparse model, fit the Cox's proportional hazards model using only predictors with indices in $\mathcal{M}^*$ by calling function ``coxph" of R package ``survival", and report the smallest absolute value of the $t$-statistic for the regression estimates. For example, for case 1, the model size is only 6 and the minimum $|t|$-statistic is computed based on these 6 estimates for each simulation. This
shows the difficulty to recover all significant variables even in
the oracle model with the minimum model size. The corresponding
boxplot for each case is shown in Figure \ref{minz}.
To demonstrate the effect of including unimportant variables, the minimum
$|t|$-statistic for the estimates of the true nonzero regression coefficients  is recalculated and shown by the boxplots in  Figure \ref{minzn}
for the model with the true important variables and 20 unimportant variables.


As expected, cases 1 and 2 are relatively easy cases, whereas cases 3 and 4 are relatively harder in the oracle model.  When we are not in the oracle model with 20 noisy variables are added, the difficulty increases as shown in Figure \ref{minzn}.  It has more impact on cases 3, 4 and 6.
%It is useful to notice that the minimum $|t|$-statistic for cases 3, 4, and 6 are smaller than that for other cases despite the much larger true coefficients due to the complex correlation structure. Even with the oracle model, it is hard to recover all important variables with such a small $|t|$-statistic.
%When 20 noisy variables are added, the difficulty increases as shown in Figure \ref{minzn}.

%{\bf Yang:  Please check the results Figures 2 and 3.  The results do not look believable.}

\section{Real data}\label{Sec:RealData}
%/home/statlab/shared/student-share/norm_NB_Cologne_502_ATlog2_grProcessedSignal/runSD.r

% In this section, we use one real data set to demonstrate the power of the proposed method. The Neuroblastoma data set is due to \cite{OBWHKSE2006}. It was used in \cite*{FanSamworthWu-09} as well for classification studies. The study includes 251 patients of the German Neuroblastoma Trials NB90-NB2004, who were diagnosed between 1989 and 2004. The patients' ages range from 0 to 296 months at diagnosis with a median age of 15 months. Neuroblastoma specimens of these 251 patients were analyzed using a customized oligonucleotide microarray. The goal is to study the association of gene expression with variable clinical information such as survival time and 3-year even free survival, among others.
%
% We obtained the neuroblastoma data from the MicroArray Quality Control phase-II (MAQC-II) project. The complete data set includes gene expression at 10,707 probe sites.  It also includes the survival information of each patient. There are five outlier arrays. After removing outlier arrays from our consideration, there are 246 patients. The survival information is available for 239 out of these 246 patients. As real data are complex, it is more appropriate to apply van-ISIS. We standard each predictor to have mean zero and standard deviation 1 and apply van-ISIS to the standardized data with $d=\lfloor  n/\log(n)\rfloor=43$. ISIS followed with SCAD penalized Cox regression selects 5 genes with probe site names:  A\_23\_P138507, A\_23\_P71319, A\_24\_P262201, A\_23\_P171034, and A\_23\_P117447.

In this section, we use one real data set to demonstrate the power of the proposed method. The Neuroblastoma data set is due to \cite{OBWHKSE2006}. It was used in \cite*{FanSamworthWu-09} for classification studies.
Neuroblastoma is an extracranial solid cancer. It is most common  in childhood and even in infancy. The annual number of incidences is  about several hundreds in the United States. Neuroblastoma is a malignant pediatric tumor originating from neural crest elements of the sympathetic nervous system.

The study includes 251 patients of the German Neuroblastoma Trials NB90-NB2004, who were diagnosed between 1989 and 2004. The patients' ages range from 0 to 296 months at diagnosis with a median age of 15 months. Neuroblastoma specimens of these 251 patients were analyzed using a customized oligonucleotide microarray.
The goal is to study the association of gene expression with variable clinical information such as survival time and 3-year event free survival, among others.

\begin{figure}[h!]
\centering{\rotatebox{0}{\includegraphics[scale=0.75]{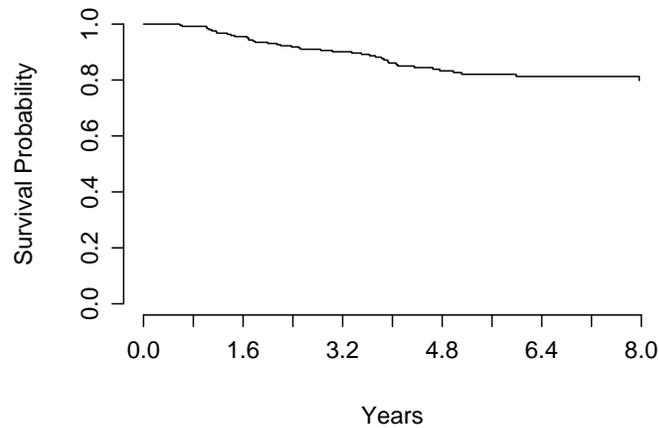}}}
\caption{Estimated survivor function for 246 patients.}\label{fig4}
\end{figure}

We obtained the neuroblastoma data from the MicroArray Quality Control phase-II (MAQC-II) project conducted by the Food Drug Administration (FDA). The complete data set includes gene expression at 10,707 probe sites.  It also includes the survival information of each patient. In this example, we focus on the overall survival.  There are five outlier arrays. After removing outlier arrays from our consideration, there are 246 patients. The (overall) survival information is available all of  these 246 patients. The censoring rate is 205/246, which is very heavy.  The survival times of those 246 patients are summarized in Figure~\ref{fig4}.

As real data are always complex, there may be some genes that are marginally unimportant but work jointly with other genes. Thus it is more appropriate to apply iterative SIS instead of SIS, since the former is more powerful. We standardize each predictor to have mean zero and standard deviation 1 and apply van-ISIS to the standardized data with $d=\lfloor  n/\log(n)\rfloor=43$. ISIS followed with SCAD penalized Cox regression selects 8 genes with probe site names: A\_23\_P31816,  A\_23\_P31816, A\_23\_P31816, A\_32\_P424973, A\_32\_P159651, Hs61272.2,    Hs13208.1,   and  Hs150167.1.

\begin{table}
\caption{Estimated coefficients for Neuroblastoma data}\label{tablenb}
 \begin{tabular}{cccc}
\hline
Probe ID & estimated coefficient & standard error & p-value\\
\hline
A\_23\_P31816& 0.864   &    0.203 &2.1e-05\\
A\_23\_P31816&-0.940   &    0.314 &2.8e-03\\
A\_23\_P31816&-0.815   &    1.704 &6.3e-01\\
A\_32\_P424973&-1.957   &    0.396 &7.8e-07\\
A\_32\_P159651&-1.295   &    0.185 &2.6e-12\\
Hs61272.2& 1.664   &    0.249 &2.3e-11\\
Hs13208.1&-0.789   &    0.149 &1.1e-07\\
Hs150167.1& 1.708   &    1.687 &3.1e-01\\
\hline
 \end{tabular}
\end{table}

\begin{figure}[h!]
\centering{\rotatebox{0}{\includegraphics[scale=0.75]{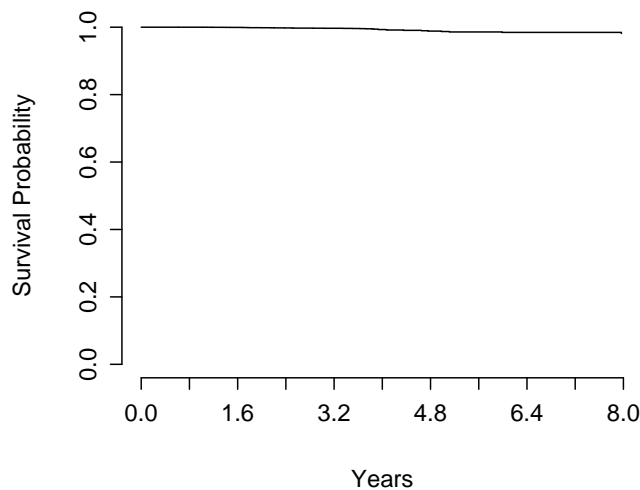}}}
\caption{Estimated baseline survivor function}\label{fignb}
\end{figure}

Now we try to provide some understanding to the significance of these selected genes in predicting the survival information in comparison to  other genes that are not selected. We first fitted the Cox's proportional hazard model with all these eight genes. Estimated coefficients are given in Table \ref{tablenb}, estimated baseline survival function is plotted in Figure \ref{fignb},  and the corresponding log-(partial)likelihood (\ref{simlik}) is -129.3517. The log-likelihood corresponding to the null model without any predictor is -215.4561. A $\chi^2$ test shows the obvious significance of the model with the eight selected  genes. Table \ref{tablenb} shows that there are two estimated coefficients that are statistically insignificant at $\alpha = 1\%$..

Next for each one of these eight genes, we remove it, fit Cox's proportional hazard model with the other seven genes, and get the corresponding log-likelihood. The eight log-likelihoods are -137.5785, -135.1846, -129.4621, -142.4066, -156.4644, -158.3799, -141.0432, and -129.8390. Their average is -141.2948, a decrease of log-likelihood by 11.9431, which is very significant with reduction one gene (the reduction of the degree of freedom by 1). In comparison to the model with the eight selected genes, $\chi^2$ tests shows significance for all selected genes except A\_23\_P31816 and Hs150167.1. This matches the p-values reported in Table \ref{tablenb}.
%It turns out that the average of these eight likelihoods is 9.233\% smaller than the likelihood corresponding to using all these eight selected genes.

Finally we randomly select 2 genes out of the genes that are not selected, fit the Cox's proportional hazard model with the above eight genes plus these two randomly selected genes, and record the corresponding log-likelihood. We repeat this process 20 times. We find that the average of these 20 new log-likelihoods is -128.3933, an increase of the log-likelihood merely by 0.9584 with two extra variables included. Comparing to the model with the eight selected genes, $\chi^2$ test shows no significance for the model corresponding to any of the 20 repetitions.

The above experiments show that the selected 8 genes are very important.  Deleting one reduces a lot of log-likelihood, while adding two random genes do not increase very much the log-likelihood.

% 0.074\% larger than the likelihood corresponding to the case using all the above eight genes. This simple simulation tells us that removing one gene from these eight genes identified by van-ISIS decreases the likelihood by 9.233\% on average. On the other hand, adding two other randomly selected genes to these eight genes only increases the likelihood by 0.074\%. It obviously shows the relative significance of these eight genes that have been selected by the van-ISIS.

%{\bf Yichao:  In addition to the analysis I suggested, I really hope that you can find the annotate of those five genes.}

\section{Conclusion}

We have developed a variable selection technique for the survival analysis with the dimensionality that can be much larger than sample size.  The focus is on the iterative sure independence screening, which iteratively applies a large-scale screening that filters unimportant variables by using the conditional marginal utility, and a moderate-scale selection by using penalized partial likelihood method, which selects further the unfiltered variables. The methodological power of the vanilla ISIS has been demonstrated via carefully designed simulation studies.  It has sure independence screening with very small false selection.  Comparing with the version of LASSO we used, it is much more computationally efficient and far more specific in selection important variables.  As a result, it has much smaller absolute deviation error and mean square error.

\end{document}